\documentclass{opt2023} 

\usepackage{comment}
\usepackage{amsmath}

\title[Optimal Transport for Kernel Gaussian Mixture]{Optimal  Transport for Kernel Gaussian Mixture Models}

\optauthor{\Name{Jung Hun Oh} \Email{ohj@mskcc.org}\\
 \addr Department of Medical Physics, 
    Memorial Sloan Kettering Cancer Center, 
    New York, NY, USA
\AND
\Name{Rena Elkin} \Email{elkinr@mskcc.org}\\
\addr Department of Medical Physics, 
    Memorial Sloan Kettering Cancer Center, 
    New York, NY, USA
\AND
\Name{Anish K. Simhal} \Email{simhala@mskcc.org}\\
\addr Department of Medical Physics, 
    Memorial Sloan Kettering Cancer Center, 
    New York, NY, USA
\AND
\Name{Jiening Zhu} \Email{jiening.zhu@stonybrook.edu}\\
\addr Department of Applied Mathematics \& Statistics,
    Stony Brook University,
    Stony Brook, NY, USA
\AND
\Name{Joseph O. Deasy} \Email{deasyj@mskcc.org}\\
\addr Department of Medical Physics, 
    Memorial Sloan Kettering Cancer Center, 
    New York, NY, USA
\AND
\Name{Allen Tannenbaum} \Email{allen.tannenbaum@stonybrook.edu}\\
\addr Departments of Computer Science and Applied Mathematics \& Statistics,
    Stony Brook University,
    Stony Brook, NY, USA
}

\begin{document}

\maketitle

\begin{abstract}%
The Wasserstein distance from optimal mass transport (OMT) is a powerful mathematical tool with numerous applications that provides a natural measure of the distance between two probability distributions. Several methods to incorporate OMT into widely used  probabilistic models, such as Gaussian or Gaussian mixture, have been developed to enhance the capability of modeling complex multimodal densities of real datasets. However, very few studies have explored the OMT problems in  a reproducing kernel Hilbert space (RKHS),  wherein  the {\it kernel trick} is utilized to avoid the need to explicitly map  input data into a high-dimensional  feature space. In the current study, we propose a Wasserstein-type metric to compute the distance between two Gaussian mixtures in a RKHS via the kernel trick, {\it i.e.}, kernel Gaussian mixture models.
\end{abstract}


\section{Introduction}\label{sec:intro}
The Gaussian mixture model (GMM) is a probabilistic model defined as a weighted sum of several Gaussian distributions \citep{Moraru2019-cx, Sanjoy_uai}. Due to  its mathematical simplicity and efficiency,  GMMs are widely used to model complex multimodal densities of real datasets \citep{Delon2019}. 

Optimal mass transport (OMT)  is an active and ever-growing field of research, originating  in the work of the French civil engineer  Gaspard Monge in 1781, which was formulated as the optimal way (via the minimization of some transportation cost) to move a pile of soil from one site to another \citep{Evans1999,Villani2003,Kantorovich2006, Pouryahya2022-cy}. OMT has made significant progress due to the pioneering effort of Leonid Kantorovich in 1942, who introduced a relaxed version of the original problem that is solved using simple linear programming  \citep{Kantorovich2006}. Recently, there has been an ever increasing growth in applications of OMT in numerous fields, including medical imaging analysis, statistical physics, machine learning, and genomics \citep{Chen2017,Chen2019,Luise2018,Zhao2013}. 

Here we briefly sketch the basic theory of OMT. Suppose that $\nu_0$ and $\nu_1$ are two absolutely continuous probability measures with compact support on $X=\mathbb{R}^d$. (The theory is valid on more general metric measure spaces.)   A Borel map $T:X\rightarrow X$ is called a \textit{transport plan} from $\nu_0$ to $\nu_1$ if it ``push-forward''  $\nu_0$ to $\nu_1$ ($T_{\#}\nu_0=\nu_1$)
which is equivalent to say that
\begin{equation}
\nu_1(B)=\nu_0(T^{-1}(B)),
\end{equation}
for every Borel subset $B\subset\mathbb{R}^d$ \citep{Kolouri2019, LEI20191}. Let $c(x,y)$ be the transportation cost to move one unit of mass from  $x$ to $y$.  The Monge version of OMT problem seeks an optimal transport map $T:X\rightarrow X$ such that the total transportation cost  $\int_{X} c(x,T(x)){\nu_0}(dx)$ is minimized over the set of all transport maps $T$. We note that the original OMT problem is highly non-linear and may not admit a viable solution. To ease this computational difficulty, Leonid Kantorovich proposed a relaxed formulation, solved by employing a linear programming method \citep{Kantorovich2006, shi_2021} which defines  the $\textit{W}_p$ Wasserstein distance between $\nu_0$ and $\nu_1$ on  $\mathbb{R}^{d}$  as follows:
\begin{equation}
W_{p}^{p} \left(  {\nu_0} , {\nu_1} \right) =
 \inf\limits_{ \pi \in \Pi (\nu_0,\nu_1)}
\int_{{\mathbb R}^{d} \times {\mathbb R}^{d}} {\Vert{ x} - { y} \Vert}^{p} d \pi(x,y),
\end{equation}
where $\Pi (\nu_0,\nu_1)$ is the set of all joint probability measures $\pi$ on $X \times X$ with $\nu_0$ and $\nu_1$ as its two marginals, and $c(x,y)$ is taken as a specific form of $c(x,y)=||x-y||^p, p\geq 1$.
In the present study, we focus on the $\textit{W}_2$ Wasserstein distance using  the squared Euclidean distance ($p=2$) as the cost function \citep{Mallasto2017}. 

While OMT ensures that the displacement interpolation (weighted barycenters) between two Gaussian distributions remains Gaussian, this property does not hold for Gaussian mixtures. To cope with this issue in GMMs, Chen ${\it et~al.}$  proposed a new Wasserstein-type distance \citep{Chen2019}. This approach optimizes the transport map between the two probability vectors of the respective Gaussian mixtures using the discrete linear program where the cost function is computed as the  closed-form formulation of the $W_2$  Wasserstein distance between Gaussian distributions. This ensures that the displacement interpolation between two Gaussian mixtures preserves the Gaussian mixture structure. Note that the sum of probabilities of all Gaussian components in a Gaussian mixture is 1, and therefore the total mass for two Gaussian mixtures is equal. 

In machine learning, kernel methods provide a powerful framework for non-linear extensions of classical linear models by  implicitly mapping the data into a high-dimensional feature space  corresponding to a reproducing kernel Hilbert space (RKHS) via a non-linear mapping
function \citep{MeantiCRR20, JMLR:v19:16-291, Oh_BMC}. 
Recently, a formulation for the $W_2$ Wasserstein distance metric between two Gaussian distributions in a RKHS  was introduced \citep{ Oh2020-KWD, Oh2020-ny, Zhang_2020}. Extending this concept, we propose an OMT framework to compute a Wasserstein-type distance between two Gaussian mixtures in  a RKHS.

\section{Methods}
In this section, we first describe  the underlying technical methods of the present work, and then introduce our proposed methodology. 

\subsection{Kernel function}
Suppose that we are given
a dataset of $n$ samples, denoted by  $X=[x_1, x_2,\cdots, x_{\it n}]\in \mathbb{R}^d$, each of which consists of $d$ features. The  data in the original input space can be mapped into a high-dimensional feature space via a non-linear mapping function $\phi$ \citep{BAUDAT2000,Oh_LDA, 1542366}: 
\begin{equation}
\Phi: X \rightarrow \mathcal{F},
\end{equation}
where $\mathcal{F}$ 
is a Hilbert space called the {\it feature space} and a (Mercer) kernel function   $k({x_i,x_j})= \langle\phi{(x_i)},\phi{(x_j)}\rangle=\phi(x_i)^{\rm T}\phi(x_j)$  is defined as an inner dot product in $\mathcal{F}$ for a positive semi-definite kernel $k:X \times X \rightarrow \mathbb{R}$ in which the kernel function is  symmetric: $k(x_i,x_j)=k(x_j,x_i)$ and $\Phi(X):=[\phi(x_1),\phi(x_2),\cdots,\phi(x_n)]$ \citep{Scholkopf2000, Rahimi2007}. The resulting  Gram matrix $K = \Phi^{\rm T}\Phi$  is positive semi-definite ($K \succeq 0$) with $K_{ij}:=k(x_i,x_j), \forall i,j \in \{1,\cdots,n\}$. This process is called the \emph{kernel trick}.
Common choices of kernel functions are the  Gaussian radial basis function (RBF) and polynomial kernels \citep{Kolouri2016, 8599461, pr8010024}. Kernels are widely used in machine learning algorithms such as  support vector machines \citep{CHEN2007161}, linear discriminant analysis \citep{Oh_LDA}, and principal component analysis \citep{Scholkopf, Sterge_2020}. In this
study, the following RBF kernel is employed: 
\begin{equation}
k({x_i,x_j})=\mathrm {exp}\big(-\gamma||{x_i-x_j}||^2 \big),\label{Eq:kernel}
\end{equation}
where $\gamma > 0$  controls the kernel width. The mean and the covariance matrix  in the feature space
are given by
\begin{equation}\label{Eq:meansigma}
\it m=\frac{\rm 1}{ n}\sum_{i={\rm 1}}^{n}\phi(x_{\it i})={\rm {\Phi} {\it s}}, ~~
{\rm \Sigma}=\frac{\rm 1}{n}\sum_{i={\rm 1}}^{n}(\phi(x_{\it
i})-{\it m})(\phi(x_{\it i})-{\it m})^{\rm T}={\rm {\Phi} {\it JJ}^{\rm T}\rm {\Phi}^{\rm T}},
\end{equation} 
where $s_{n \times 1}=\frac{1}{n}{\vec{1}}^{\rm T}$, $ {J}=\frac{\rm 1}{\sqrt{\it {n}}}(I_{\it n}-s{\vec{1}})$, 
${\vec{1}}=[1,1,\cdots,1]$, and $I_n$ is the $n \times n$ identity matrix. Then, denoting ${\Phi J}$ by $S$, we have
\begin{equation}
 {S=\Phi J} = \frac{\rm 1}{\sqrt{\it n}}[(\phi({ x}_{\rm 1})-\it m),
\cdots, (\phi({ x}_{\it n})-\it m)].\label{Eq:W}
\end{equation}
A key idea to compute a Wasserstein-type distance between two Gaussian
mixtures in a RKHS is to efficiently convert the mapping functions represented by $\it m$ and $\Sigma$ to kernel functions by putting two mapping functions next to each other in a certain way. 

\subsection{OMT between Gaussian distributions}
Given two Gaussian distributions, $\nu_0$ and $\nu_1 \in \mathbb{R}^{\it
d}$, with mean $\theta_i$ and covariance matrix $\it C_i$ for $i=0$ and 1, the $W_2$ Wasserstein distance between the two distributions has the following closed formula:
\begin{equation}
W_2 ( \nu _{0},  \nu _{1} ) ^{2}
  =   \Vert{\theta}_{\rm 0}- {\theta} _{\rm 1} \Vert  ^{2}+\\
 \mathrm {tr} \left({\it C}_0+ {\it C}_1 -2 \left(   {\it C}_0^{\frac{1}{2}}  {\it C}_1  {\it C}_0^{\frac{1}{2}} \right)^{\frac{1}{2}} \right),\label{Eq:w2}
\end{equation}
 where tr is the trace and $\mathrm {tr} \left({\it C}_0+ {\it C}_1 -2 (   {\it C}_0^{\frac{1}{2}}  {\it C}_1  {\it C}_0^{\frac{1}{2}} )^{\frac{1}{2}} \right)=\mathrm {tr} \left({\it C}_0+ {\it C}_1 -2 (    {\it C}_1  {\it C}_0 )^{\frac{1}{2}} \right)$. In particular, when ${\it C}  _{0}={\it C}  _{1}$, we have $W_2 ( \nu _{0},  \nu _{1} )^{2}
  =   \Vert{\theta}_{\rm 0}- {\theta} _{\rm 1} \Vert^{2}.$
  If ${\it C_0}$  and ${\it C_1}$ are non-degenerate, the geodesic path (displacement interpolation)  $(\nu_t)_{t\in[0,1]}$ between $\nu_0$ and $\nu_1$ remains Gaussian and satisfies 
  \begin{equation}
\nu_t \in {\rm argmin}_\rho (1-t)W_2(\nu_0,\rho)^2+tW_2(\nu_1, \rho)^2,
 \end{equation}
 with  mean $\theta_t=(1-t)\theta_{\rm 0}+t{\theta} _{\rm 1}$ and  covariance matrix given by:
\begin{equation}
C_t=((1-t)I_d+tQ)C_0((1-t)I_d+tQ),
\end{equation}
where $I_d$ is the $d \times d$ identity  matrix and $Q=C_1^{\frac{1}{2}}(C_1^{\frac{1}{2}}C_0C_1^{\frac{1}{2}})^{-{\frac{1}{2}}}C_1^{\frac{1}{2}}$ \citep{Delon2019}.

\subsection{OMT between Gaussian distributions in a RKHS}
The data distribution in a RKHS represented via proper kernel functions is assumed to approximately follow a Gaussian distribution under suitable conditions, as justified by Huang  ${\it et~al.}$ \citep{Huang2005}. Let $\rho_{0}$ and $\rho_{1}  \in \mathbb{R}^{\it
l}$ be two Gaussian distributions in a RKHS with mean ${\it m}_{\it i}$ and covariance matrix ${\Sigma}_{\it i}$ for $i=0,1$. The $W_{2}$ Wasserstein distance in a RKHS, denoted as $KW_2$ ({\em kernel Wasserstein distance}), is then defined as follows:
\begin{equation}
KW_{2} ( \rho _{0},  \rho _{1} ) ^{2}
  =   \Vert{\it m}_{\rm 0}- {\it m} _{\rm 1} \Vert  ^{2} +\\
  \mathrm {tr} \left({\Sigma}_0+  \Sigma_1 -2 \left(   {\Sigma}_0^{\frac{1}{2}}  {\Sigma}_1  {\Sigma}_0^{\frac{1}{2}} \right) ^{\frac{1}{2}} \right),\label{Eq:kw2}
\end{equation}
where $\Sigma_i^{\frac{1}{2}}$ is the unique semi-definite positive square root of a symmetric semi-definite positive matrix $\Sigma_i$ \citep{Xia, DelonAgnes_2022}.

Suppose that there are two sets of  data  in the original input space, ${X}=[{x}_1, {x}_2,\cdots, {x}_{\it n}]$ and  ${Y}=[{y}_1, {y}_2,\cdots, {y}_{\it m}] \in \mathbb{R}^d~ (d < l)$, associated with $\rho_{0}$ and $\rho_{1}$, respectively.
The first term in Eq.~(\ref{Eq:kw2}) is the squared maximum mean
discrepancy (MMD) \citep{pmlr-v32-iyer14, Ouyang_mmd} and may be expressed with kernel functions as follows: 
\begin{equation}
\Vert {\it m} _{0} - {\it m} _{1} \Vert  ^{2}=\frac{1}{n^{2}} \sum _{i=1}^{n} \sum _{j=1}^{n}{\it k} \left( {x}_{i},{x}_{j} \right) - 
  \frac{2}{nm} \sum _{i=1}^{n} \sum _{j=1}^{m}{\it k} \left( {x}_{i},{y}_{j} \right)
 + \frac{1}{m^{2}} \sum _{i=1}^{m} \sum _{j=1}^{m}{\it k} \left( {y}_{i},{y}_{j} \right).\label{Eq:firstterm}
\end{equation}
By using  Eq.~(\ref{Eq:meansigma}), the second term in Eq.~(\ref{Eq:kw2}) is expressed as follows:
\begin{eqnarray}
\mathrm{tr} \left( {\Sigma}_0+  \Sigma_1-2 \left(   {\Sigma}_0^{\frac{1}{2}}  {\Sigma}_1  {\Sigma}_0^{\frac{1}{2}} \right) ^{\frac{1}{2}} \right)
  =\mathrm{tr} \left( {\Sigma}_0+  \Sigma_1-2 \left(   {\Sigma}_1 {\Sigma}_0 \right) ^{\frac{1}{2}} \right)~~~~~~~~~~ \\\nonumber
 =\mathrm{tr} \left(J_0J_0^{\rm T}{\Phi}_0^{\rm T}{\Phi}_0\right)+\mathrm{tr} \left(J_1J_1^{\rm T}{\Phi}_1^{\rm T}{\Phi}_1\right)- 2\mathrm{tr}\left({\Phi}_1 J_1J_1^{\rm T}K_{10}J_0J_0^{\rm T}{\Phi}_0^{\rm T}\right)^{\frac{1}{2}}
 \\\nonumber
 =\mathrm{tr} \left( {J}_{0}{J}_{0}^{\rm T}{K}_{00} \right) +\mathrm{tr} \left( {J}_{1}{J}_{1}^{\rm T}{{K}_{11}} \right) -2 \mathrm{tr} \left(  {\Phi} _{1}{Q} {\Phi} _{0}^{\rm T} \right) ^{\frac{1}{2}},~~~~~~~~~~~~~~~~~~~~~~~~~~\label{Eq:secondterm}
\end{eqnarray}
where
$\mathrm{tr} \left(  {\Phi} _{1}{Q} {\Phi} _{0}^{\rm T} \right) ^{\frac{1}{2}}=\mathrm{tr} \left( {K}_{01}{J}_{1}{J}_{1}^{\rm T}{K}_{10}{J}_{0}{J}_{0}^{\rm T} \right) ^{\frac{1}{2}}$, ${Q}={J}_{1}{J}_{1}^{\rm T}{K}_{10}{J}_{0}{J}_{0}^{\rm T}$,  and ${K} _{\it ij}= {\Phi} _{i}^{\rm T}{\Phi_{\it j}}$. 
Note that  ${\Sigma_{\rm 0}}$ and ${\Sigma_{\rm 1}}$ are symmetric positive semi-definite. 
Plugging the two terms together into Eq.~(\ref{Eq:kw2}), the ${\it KW_{\rm 2}}$  distance between $\rho_{0}$ and $\rho_{1}$ may be expressed as:
\begin{eqnarray}
KW_2(\rho _{0},  \rho _{1} ) ^{2} =   \frac{1}{n^{2}} \sum _{i=1}^{n} \sum _{j=1}^{n}{\it k} \left( {x}_{i},{x}_{j} \right) -    
\frac{2}{nm} \sum _{i=1}^{n} \sum _{j=1}^{m}{\it k} \left( {x}_{i},{y}_{j} \right)
 +\frac{1}{m^{2}} \sum _{i=1}^{m} \sum _{j=1}^{m}{\it k} \left( {y}_{i},{y}_{j} \right)+ \\\nonumber
\mathrm{tr} \left( {J}_{0}{J}_{0}^{\rm T}{K}_{00} \right)+\mathrm{tr} \left( {J}_{1}{J}_{1}^{\rm T}{K}_{11} \right) -  
   2\mathrm{tr} \left( {K}_{01}{J}_{1}{J}_{1}^{\rm T}{K}_{10}{J}_{0}{J}_{0}^{\rm T} \right) ^{\frac{1}{2}}.   \label{Eq:final_kwd}
\end{eqnarray}
Eq.~(13) will be used as a key component for the computation of a Wasserstein-type distance between two Gaussian
mixtures in a RKHS in the following section. 
In the special case, when ${\Sigma}  _{0}={\Sigma}  _{1}$, we have $KW_2(\rho _{0},  \rho _{1} )^{2}=\frac{1}{n^{2}} \sum _{i=1}^{n} \sum _{j=1}^{n}{\it k} \left( {x}_{i},{x}_{j} \right) - 
  \frac{2}{nm} \sum_{i=1}^{n} \sum_{j=1}^{m}{\it k} \left({x}_{i},{y}_{j}\right)
 +\\ \frac{1}{m^{2}} \sum _{i=1}^{m} \sum _{j=1}^{m}{\it k} \left( {y}_{i},{y}_{j} \right)$.

\subsection{OMT between GMMs in a RKHS}
Based on the OMT method between GMMs introduced in \citep{Chen2019}, we propose a Wasserstein-type metric to compute the
distance between two Gaussian mixtures in a RKHS via the kernel trick, which preserves the Gaussian mixture structure in the displacement interpolation. Let $\mu$ be a Gaussian mixture  in a RKHS: 
\begin{eqnarray}
\mu=\sum_{k=1}^{N}p^{k}v^{k},
\end{eqnarray}
where each $v^k$ is a Gaussian distribution in a RKHS with $v^{k}=\mathcal{N}(m_k, \Sigma_k)$ and $p^k$ is a probability of $v^k$ with $\sum_{k=1}^{N}p^{k}=1$.
Let $\mu_0$ and $\mu_1$ denote two Gaussian mixtures  in a RKHS in the following form:
\begin{eqnarray}
\mu_i=p_i^1v_i^1+p_i^2v_i^2+...+p_i^{N_i}v_i^{N_i}, \quad i=0,1,
\end{eqnarray}
where $N_i$ is the number of Gaussian components of $\mu_i$. The distance between $\mu_0$ and $\mu_1$ is then defined according to the  discrete OMT formulation for discrete measures \citep{Chen2019, Mathews2020-be}:
\begin{eqnarray}
d(\mu_0,\mu_1)^2=\min\limits_{ \pi \in \Pi (p_0,p_1)}\sum_{i,j}c_{ij}\pi_{ij},
\label{Eq:dis}
\end{eqnarray}
where $\Pi (p_0,p_1)$  is the set of all joint probability measures between $p_0$ and $p_1$, defined as:
\begin{eqnarray}\nonumber
\Pi (p_0,p_1)=\{\pi \in \mathbb{R}_{+}^{N_0 \times N_1} | \sum_{j}\pi_{kj}=p_0^k,~ \sum_{k}\pi_{kj}=p_1^j\}. 
\end{eqnarray}
The cost $c_{ij}$ is taken to be the square of the $KW_2$  distance between $v_0^i$ and $v_1^j$ in a RKHS:
\begin{eqnarray}
c_{ij}=KW_2(v_0^i, v_1^j)^2.
\end{eqnarray}
Since $v_0^i$ and $v_1^j$ are Gaussian distributions in a RKHS, the $KW_2$ distance can be computed using Eq. (13). Let $\pi^*$ be the optimal solution of Eq.~(\ref{Eq:dis}). Then, the distance $d(\mu_0,\mu_1)$ between  two  Gaussian mixtures in a RKHS is defined as follows:
\begin{eqnarray}
d(\mu_0,\mu_1)=\sqrt{\sum_{i,j}c_{ij}\pi_{ij}^*},
\end{eqnarray}
where $d(\mu_0, \mu_1) \geq KW_2(\mu_0, \mu_1)$ and the following property holds \citep{Chen2019}:
\begin{eqnarray}
d(\mu_s,\mu_t)=(t-s)d(\mu_0, \mu_1), \quad 0 \leq s < t \leq 1.
\end{eqnarray}
The geodesic path $\mu_t$ between $\mu_0$ and $\mu_1$ is defined as
\begin{eqnarray}
\mu_t=\sum_{i,j}\pi_{ij}^*v_t^{ij},
\end{eqnarray}
where $v_t^{ij}$ is the displacement interpolation between two Gaussian distributions, $v_0^i$ and $v_1^j$.

\section{Experiments}
\begin{figure}[ht]
\includegraphics[width=0.36\textwidth]{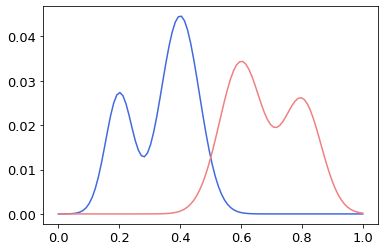}
\centering
	\caption{As an example, two Gaussian mixtures were created with  $\mu_0=0.3\mathcal{N}(0.2,0.002)+0.7\mathcal{N}(0.4,0.004)$ in blue and $\mu_1=0.6\mathcal{N}(0.6,0.005)+0.4\mathcal{N}(0.8,0.004)$ in red, in a 1-dimensional space. }
	\label{fig1}
\end{figure}

\begin{figure*}[ht]
	\centering
\includegraphics[width=0.95\textwidth]{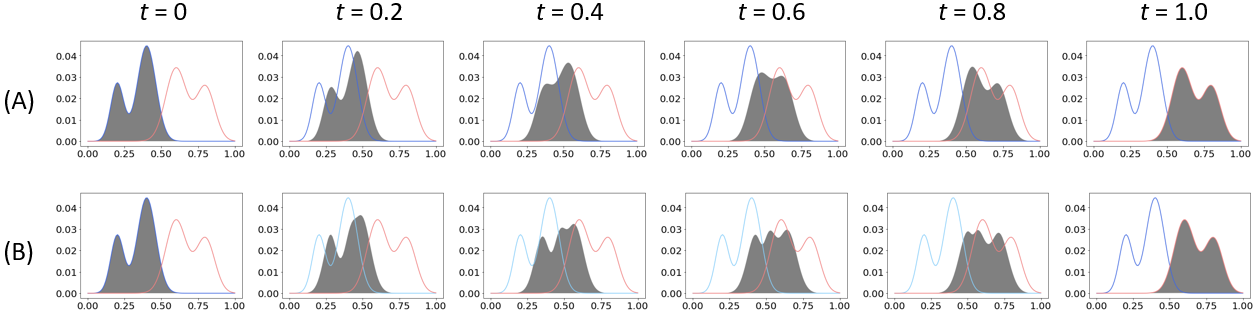}
	\caption{Displacement
interpolation (gray) $\mu_t$ between two Gaussian mixtures, $\mu_0$  (blue curve) and $\mu_1$  (red curve).
(A) Displacement
interpolation for the metric $d(\mu_0,\mu_1)$ and (B)    general Wasserstein distance.}
	\label{fig2}
\end{figure*}

Suppose that  we are given two Gaussian mixtures, $\mu_0=0.3\mathcal{N}(0.2,0.002)+0.7\mathcal{N}(0.4,0.004)$ and $\mu_1=0.6\mathcal{N}(0.6,0.005)+0.4\mathcal{N}(0.8,0.004)$ as a  1-dimensional example, as shown in Figure~1\vphantom{\ref{fig1}}.
Figure~2\vphantom{\ref{fig2}} shows the displacement
interpolation $\mu_t$ for both the metric $d(\mu_0,\mu_1)$ and  general Wasserstein distance  between the
$\mu_0$ and $\mu_1$ at $t$ = 0, 0.2, 0.4, 0.6,  0.8, and 1.0. It is observed that the displacement
interpolation for the metric $d(\mu_0,\mu_1)$ preserves the Gaussian mixture structure, whereas the displacement
interpolation for the general Wasserstein distance does not.

\begin{figure*}[ht]
	\centering
\includegraphics[width=1\textwidth]{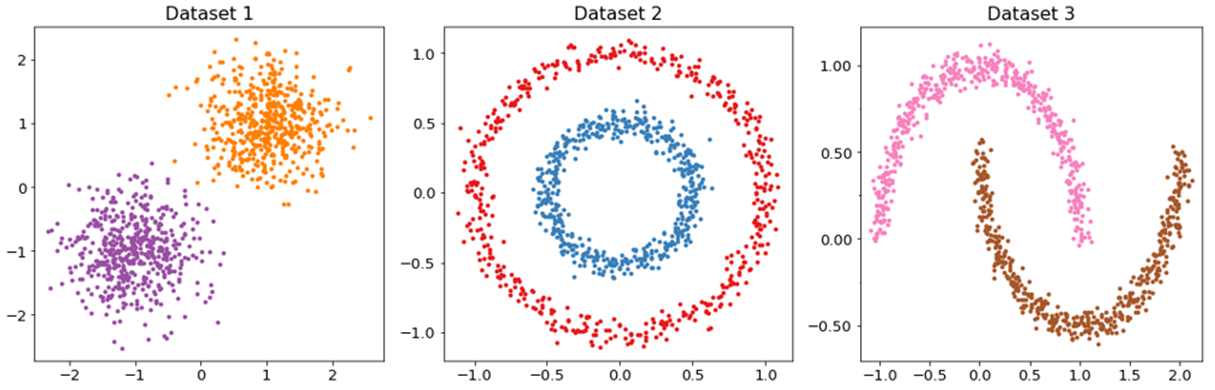}
	\caption{Three datasets,  each of which has two distributions, which were used for Gaussian mixture analysis in a RKHS. We assume that each dataset follows a Gaussian mixture, {\it i.e.}, $\mu_0 = p_0^1v_0^1 + p_0^2v_0^2$, $\mu_1 = p_1^1v_1^1 + p_1^2v_1^2$, and $\mu_2 = p_2^1v_2^1 + p_2^2v_2^2$, respectively, in a RKHS. }
	\label{fig3}
\end{figure*}

As another example, three datasets were generated in the original input space, each of which has two distributions (Figure~3\vphantom{\ref{fig3}}). Each dataset consists of 1,000 data points with 500 data points for each distribution. The $d(\cdot, \cdot)$ was then computed between each pair of datasets in a RKHS, assuming that each dataset ($\mu_0, \mu_1$, and $\mu_2$) has two Gaussian components with $v_0^1$ (purple)/$v_0^2$ (orange), $v_1^1$ (red)/$v_1^2$ (blue), and $v_2^1$ (pink)/$v_2^2$ (brown), respectively in a RKHS (Tables 1 and 2). In the current study, $\gamma = 1$ (Table 1) and $\gamma = 10$ (Table 2)  were used in the RBF kernel shown in Eq.~(\ref{Eq:kernel}).

The $d(\cdot, \cdot)$ values with $\gamma=10$ were larger than those corresponding to $\gamma=1$. Overall, the $d(\cdot, \cdot)$ values between dataset 2 and dataset 3 were smaller than dataset 1 vs. dataset 2 and dataset 1 vs. dataset 3.   When $\gamma=1$ and $(p_0^1, p_0^2)$ is (0.1, 0.9) and (0.3, 0.7),  the $d(\cdot, \cdot)$ values  between dataset 1 and dataset 2 were larger than those between dataset 1 and dataset 3. By contrast, when $(p_0^1, p_0^2)$ is one of (0.5, 0.5), (0.7, 0.3), and (0.9, 0.1), the $d(\cdot, \cdot)$ values  between dataset 1 and dataset 2 were smaller than those between dataset 1 and dataset 3. When $\gamma=10$,  the $d(\cdot, \cdot)$ values  between dataset 1 and dataset 3 were larger than those between dataset 1 and dataset 2 in more cases compared to when $\gamma=1$.

\begin{figure*}[ht]
	\centering
\includegraphics[width=1\textwidth]{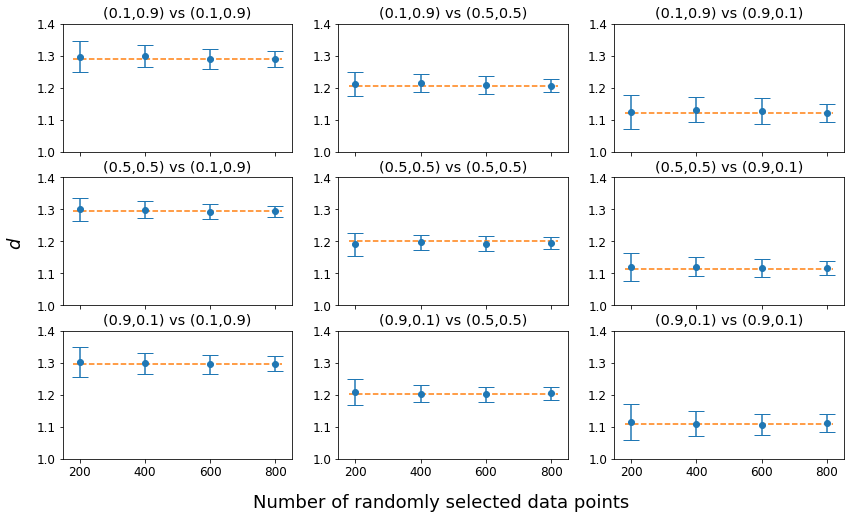}
\caption{Using dataset 1 and dataset 2, simulation tests were conducted with (0.1, 0.9), (0.5, 0.5), and (0.9, 0.1) for both $(p_0^1, p_0^2)$ and $(p_1^1, p_1^2)$ by randomly sampling 200, 400, 600, and 800 data points from each dataset. The blue dot and error bar indicate the average distance and standard deviation after 100 repetitions  of each sampling experiment with the combination of $(p_0^1, p_0^2)$ and $(p_1^1, p_1^2)$. The horizontal dot line indicates  $d(\cdot, \cdot)$ when the original data with 1,000 data points for each dataset were tested. }
	\label{fig4}
\end{figure*}

Using dataset 1 and dataset 2, simulation tests were conducted with (0.1, 0.9), (0.5, 0.5), and (0.9, 0.1) for both $(p_0^1, p_0^2)$ and $(p_1^1, p_1^2)$ by randomly sampling 200, 400, 600, and 800 data points from 1,000 data points of each dataset. For each sampling experiment of the combination of $(p_0^1, p_0^2)$ and $(p_1^1, p_1^2)$, 100 tests were conducted and  the average distance and standard deviation were computed. In Figure~4\vphantom{\ref{fig4}}, the horizontal dot line indicates  $d(\cdot, \cdot)$ when the original data with 1,000 data points for each dataset were analyzed. As can be seen, as the number of randomly selected data points increases, the standard deviation becomes narrower, converging to the horizontal dot line. Not surprisingly, the average distance after 100 repetitions  of each sampling experiment was very similar to  $d(\cdot, \cdot)$ computed on the original data with 1,000 data points for each dataset.

\begin{figure}[ht]
\centering
\includegraphics[width=0.49\textwidth]{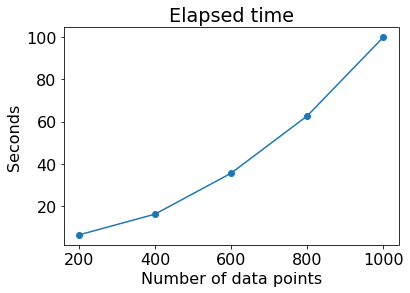}
	\caption{Elapsed time to compute  $d(\cdot, \cdot)$ between dataset 1 and dataset 2 with randomly sampled data points (200, 400, 600, 800) from each dataset, compared to the elapsed time in the original datasets (each 1,000 data points). }
	\label{fig5}
\end{figure}

Figure~5\vphantom{\ref{fig5}} illustrates the elapsed time to compute  $d(\cdot, \cdot)$ between dataset 1 and dataset 2 when each test was conducted only once with randomly sampled data points (200, 400, 600, 800) from each dataset, compared to the elapsed time in the original datasets  (each 1,000 data points). As the number of data points increased, the computational time to compute  $d(\cdot, \cdot)$ sharply increased. Therefore, there is a tradeoff between the computational time and the precision of computed distance in the sampling approach. Use of advanced sampling techniques will help compute the distance within reasonable computational time and with minimal error.
All experiments in the current study were  carried out using Python language in Google Colab-Pro environment.

\acks{This study was supported in part by National Institutes of Health (NIH)/National Cancer Institute Cancer Center support grant (P30 CA008748), NIH grant (R01-AG048769), AFOSR grant (FA9550-20-1-0029), Army Research Office grant  (W911NF2210292),  Breast Cancer Research Foundation grant (BCRF-17-193), and a grant from the Cure Alzheimer's Foundation. }

\newpage
\bibliography{OPT_2023_Final}
\newpage
\clearpage
\appendix

\section{Related work}
The Gaussian model has numerous applications in data analysis due to its  mathematical tractability and simplicity. In particular, a closed-form formulation of $W_2$ Wasserstein distance for  Gaussian densities allows for the extension of applications of OMT in connection with Gaussian processes. 
As a further extension,   Janati  ${\it et~al.}$ proposed   an entropy-regularized OMT  method between two Gaussian measures, by solving the fixed-point
equation underpinning the Sinkhorn algorithm for both the balanced and unbalanced cases \citep{Janati, Cuturi_sinkhorn, e23030302}. Mallasto ${\it et~al.}$ introduced an alternative approach for the entropy-regularized optimal transport, providing closed-form expressions and interpolations between Gaussian measures \citep{Masarotto2022}. Le  ${\it et~al.}$ investigated the entropic  Gromov-Wasserstein distance
between (unbalanced) Gaussian distributions and  the entropic Gromov-Wasserstein barycenter of multiple Gaussian distributions \citep{pmlr-v162-le22a, Jin_uai, Jin_IJCAI}.

Kernel methods are extensively employed in machine learning, providing a powerful capability to efficiently handle data in a non-linear space by implicitly mapping data into a high-dimensional  space, a method known as the {\it kernel trick}. Ghojogh  ${\it et~al.}$ comprehensively reviewed the background theory of kernels and their applications  in machine learning \citep{Ghojogh}. As an effort to incorporate kernels into OMT, Zhang ${\it et~al.}$ proposed a solution to compute the $W_2$ Wasserstein distance  between Gaussian measures in a RKHS \citep{Zhang_2020} and Oh ${\it et~al.}$ proposed an alternative  algorithm called the {\it kernel Wasserstein distance}, giving an explicit detailed proof \citep{Oh2020-ny}.  Minh introduced a novel algorithm to compute the entropy-regularized $W_2$ Wasserstein distance between Gaussian measures in a RKHS  via the finite kernel Gram matrices, providing explicit closed-form formulas along with the Sinkhorn barycenter equation with a unique non-trivial solution \citep{Minh_2022}.

Numerous studies have been conducted on GMMs in connection with other techniques. Wang ${\it et~al.}$ proposed a kernel trick embedded GMM method by employing the Expectation  Maximization (EM) algorithm  to deduce a  parameter estimation method for GMMs in the feature space, and introduced a Monte Carlo sampling technique to speed up the computation in large-scale data problems \citep{Wang_kernel, Graca_neurips, 9416576}. 
Chen ${\it et~al.}$ proposed a new algorithm to compute a Wasserstein-type distance between two Gaussian mixtures, employing the closed-form solution between Gaussian measures as the cost function,  represented as the discrete optimization problem \citep{Chen2019}. Following this latter work, Delon and Desolneux investigated the theory of the  Wasserstein-type distance on GMMs, showing several applications including color transfer between images  \citep{Delon2019}. Mathews ${\it et~al.}$ applied the GMM Wasserstein-type distance to functional network analysis utilizing RNA-Seq gene expression profiles from The Cancer Genome Atlas (TCGA) \citep{Mathews2020-be}. This approach enabled the identification of gene modules (communities) based on the local connection structure of the gene network and the collection of joint distributions of nodal  neighborhoods.  However, to date, no study has explored the incorporation of the kernel trick embedded GMM into OMT. To address this, we propose an OMT solution to compute the distance between Gaussian mixtures in a RKHS.

\section{Entropy-regularized optimal transport between
Gaussians }

Given $\mu, \nu \in \mathcal{P}$ and the cost function $c$, the entropic optimal transport (OT) problem proposed by Mallasto {\it et al.} is expressed as follows \citep{Masarotto2022}: 
\begin{equation}
{\rm OT}^\epsilon(\mu, \nu)= {\rm min}_{\gamma \in \pi(\mu,\nu)}\big\{{\mathbb{E}_\gamma[c]+\epsilon D_{\rm KL}(\gamma||\mu \otimes \nu)}\big\},
\end{equation}
which relaxes the OT problem employing a Kullback-Leibler (KL)-divergence term, yielding a strictly convex problem for $\epsilon > 0$.  Let $\nu_0$ and $\nu_1 \in \mathbb{R}^{\it
d}$ be two Gaussian distributions with mean $\theta_i$ and covariance matrix $\it C_i$ for $i=0$ and 1. Then, the entropy-regularized $W_2$ Wasserstein distance can be computed using a closed-form defined as:
\begin{eqnarray}
W_2^\epsilon(\nu_0, \nu_1)^2=\Vert{\theta}_{\rm 0}- {\theta} _{\rm 1} \Vert  ^{2} + \rm {tr}({\it C}_0)+ \rm {tr}({\it C}_1) - \mathcal{B},\\\nonumber
\mathcal{B}=\frac{\epsilon}{2}\left({\rm tr}(\it {M}^\epsilon) -\rm{log~}\rm{det}({\rm}{\it M}^\epsilon)+\it {d}\rm{log}2-2\it {d}\right),\label{ent1}
\end{eqnarray}
where $M_{ij}^\epsilon=I+(I+\frac{16}{\epsilon^2}{\it C}_i C_j)^{1/2}$, assuming
\begin{eqnarray}
\gamma=\mathcal{N}(0, \Gamma), ~~\Gamma=
\begin{bmatrix}
C_0 & G^{\rm T}\\
G & C_1
\end{bmatrix}.
\end{eqnarray}
The entropic  displacement interpolation between $\nu_0$ and $\nu_1$ also follows Gaussian as $\nu_t = \mathcal{N}(\theta_t, C_t)$ for $t \in[0,1]$ with $\theta_t=t\nu_0+(1-t)\nu_1$ and 
\begin{equation}
C_t = (1-t)^2C_0+t^2C_1+t(1-t)\left(\left(\frac{\epsilon^2}{16}I+C_0C_1\right)^{1/2}+\left(\frac{\epsilon^2}{16}I+C_1C_0\right)^{1/2}\right).
\end{equation}
The entropic barycenter for  $N$ distributions with $\nu_0, \nu_1, \cdots, \nu_{N-1}$ is expressed as follows: 
\begin{equation}
\theta^*=\sum_{i=1}^N\lambda_i \theta_i,~~ C^*=\frac{\epsilon}{4}\sum_{i=1}^N \lambda_i \left(-I+\left(I+\frac{16}{\epsilon^2}C^{*1/2}C_iC^{*1/2}\right)^{1/2}\right).
\end{equation}

Similarly, Janati {\it et al.} defined a closed-form for the entropy-regularized $W_2$ Wasserstein distance between Gaussians as follows \citep{Janati}: 
\begin{eqnarray}
W_2^\sigma(\nu_0, \nu_1)^2=\Vert{\theta}_{\rm 0}- {\theta} _{\rm 1} \Vert^{2} + \rm {tr}({\it C}_0)+ \rm tr ({\it C}_1) - \mathcal{F},\\\nonumber
\mathcal{F}={\rm tr}(D_\sigma)-d\sigma^2(1-{\rm log}(2\sigma^2))-\sigma^2{\rm log~det}(D_\sigma+\sigma^2I),
\end{eqnarray}
where $D_\sigma=(4C_0C_1+\sigma^4I)^{1/2}$ and $\sigma>0$.

\section{Entropy-regularized optimal transport between
Gaussians in RKHS }
Here we propose two new formulas to solve the entropy-regularized $W_2$ Wasserstein distance between Gaussians in RKHS using kernel trick.

\textbf {Proposition 1.} Let $\rho_{i}=\mathcal{N}(m_i, \Sigma_i)$ for $i=0, 1$  be two Gaussian distributions  on $\mathbb{R}^l$ in RKHS, and let two sets of  data  in the input space be ${X}=[{x}_1, {x}_2,\cdots, {x}_{\it n}]$ and  ${Y}=[{y}_1, {y}_2,\cdots, {y}_{\it m}] \in \mathbb{R}^d ~(l > d)$ associated with $\rho_{0}$ and $\rho_{1}$, respectively. Then,  a closed-form solution for Eq.~(22) in RKHS exists, denoted as $KW_{2}^\epsilon(\rho_0, \rho_1)^2$:
\begin{eqnarray}
KW_{2}^\epsilon(\rho_0, \rho_1)^2={\Vert{m}_{\rm 0}- {m} _{\rm 1} \Vert  ^{2}} + {\rm {tr}(\Sigma_0)+ \rm {tr}(\Sigma_1)} - {\mathcal{B}},\\\nonumber
\mathcal{B}=\frac{\epsilon}{2}\left({\rm tr}(\it {M}^\epsilon) -\rm{log~}\rm{det}({\rm}{\it M}^\epsilon)+\it {l}\rm{log}2-2\it {l}\right).
\end{eqnarray}
{\bf Proof.} Let $A=\Sigma_0\Sigma_1$. The trace of a square matrix $A$ is defined as the trace of the eigenvalue matrix of $A$, i.e., ${\rm tr}(A)={\rm tr}(\Lambda)$ in $AP=P\Lambda$  where $P$ and $\Lambda$ are the
estimated eigenvector and eigenvalue matrices, respectively.   Let $\lambda_{1}, \lambda_{2}, \cdots, \lambda_{k} > 0$ be the distinct eigenvalues of  $A$. 
Then, the eigenvalues of $(I+\frac{16}{\epsilon^2}A)^{1/2}$  are $\sqrt{1+\frac{16}{\epsilon^2}\lambda_{1}}, \sqrt{1+\frac{16}{\epsilon^2}\lambda_{2}}, \cdots,\sqrt{1+\frac{16}{\epsilon^2}\lambda_{k}}$ and

\begin{equation}
{\rm tr}(M^\epsilon)={\rm tr}\left(I+\left(I+\frac{16}{\epsilon^2}\Sigma_0\Sigma_1\right)^{1/2}\right)=l+\sum_{i=1}^k \sqrt{1+\frac{16}{\epsilon^2}\lambda_i}.
\end{equation}
By  ${\rm{det}}(I+tA)=(1+t\lambda_1)(1+t\lambda_2)\cdots(1+t\lambda_k)$, we compute ${\rm log~det}(M^\epsilon)$ as follows:
\begin{eqnarray}
{\rm log~det}(M^\epsilon)&=&{\rm log~ det}\left(I+\left(I+\frac{16}{\epsilon^2}\Sigma_0\Sigma_1\right)^{1/2}\right) \\\nonumber
&=&{\rm log}\left(1+\sqrt{1+\frac{16}{\epsilon^2}\lambda_{1}}\right)\left(1+\sqrt{1+\frac{16}{\epsilon^2}\lambda_{2}}\right)\cdots\left(1+\sqrt{1+\frac{16}{\epsilon^2}\lambda_{k}}\right)\\\nonumber
&=&\sum_{i=1}^k {\rm log}\left(1+\sqrt{1+\frac{16}{\epsilon^2}\lambda_{i}}\right).
\end{eqnarray}
Therefore, $\mathcal{B}$ is expressed as:
\begin{eqnarray}
\mathcal{B}&=&\frac{\epsilon}{2}\left({\rm tr}(\it {M}^\epsilon) -\rm{log~}\rm{det}({\rm}{\it M}^\epsilon)+\it {l}\rm{log}2-2\it {l}\right)\\\nonumber
&=&\frac{\epsilon}{2}\left(\sum_{i=1}^k \sqrt{1+\frac{16}{\epsilon^2}\lambda_i} - \sum_{i=1}^k {\rm log}\left(1+\sqrt{1+\frac{16}{\epsilon^2}\lambda_{i}}\right) - {\it l}{\rm log5}\right).
\end{eqnarray}
Finally, we have a closed-form solution for the entropy-regularized $W_2$ Wasserstein distance between Gaussians in RKHS:
\begin{eqnarray}
KW_{2}^\epsilon(\rho_0, \rho_1)^2=\frac{1}{n^{2}} \sum _{i=1}^{n} \sum _{j=1}^{n}{\it k} \left( {x}_{i},{x}_{j} \right) -    
\frac{2}{nm} \sum _{i=1}^{n} \sum _{j=1}^{m}{\it k} \left( {x}_{i},{y}_{j} \right)
 +\frac{1}{m^{2}} \sum _{i=1}^{m} \sum _{j=1}^{m}{\it k} \left( {y}_{i},{y}_{j} \right)+~\\\nonumber
 \mathrm{tr} \left( J_{0}J_{0}^{\rm T}K_{00} \right) +\mathrm{tr} \left( J_{1}J_{1}^{\rm T}{K_{11}} \right) 
 -\frac{\epsilon}{2}\left(\sum_{i=1}^k \sqrt{1+\frac{16}{\epsilon^2}\lambda_i} - \sum_{i=1}^k {\rm log}\left(1+\sqrt{1+\frac{16}{\epsilon^2}\lambda_{i}}\right) -{\it l}{\rm log5}\right).
\end{eqnarray}

\textbf {Proposition 2.} 
Similarly, we have a closed-form solution for Eq.~(26) in RKHS. \\
{\bf Proof.} Since the first two terms in Eq. (26) are the same as those in Eq. (22), we only solve  $\mathcal{F}={\rm tr}(D_\sigma)-l\sigma^2(1-{\rm log}(2\sigma^2))-\sigma^2{\rm log~det}(D_\sigma+\sigma^2I)$ in RKHS. As in Proposition 1, let $\lambda_{1}, \lambda_{2}, \cdots, \lambda_{k} > 0$ be the distinct eigenvalues of  $\Sigma_0\Sigma_1$ that can be computed using $\mathrm{tr}(\Sigma_0\Sigma_1)=\mathrm{ tr} \left( K_{10}J_{0}J_{0}^{\rm T}K_{01}J_{1}J_{1}^{\rm T} \right)$. Then, the eigenvalues of $D_\sigma=(4\Sigma_0\Sigma_1+\sigma^4I)^{1/2}=\sigma^2(\frac{4}{\sigma^4}\Sigma_0\Sigma_1+I)^{1/2}$  are $\sigma^2\sqrt{\frac{4}{\sigma^4}\lambda_1+1}, \sigma^2\sqrt{\frac{4}{\sigma^4}\lambda_2+1}, \cdots, \sigma^2\sqrt{\frac{4}{\sigma^4}\lambda_k+1}$. 
Therefore, $\mathrm{tr}(D_\sigma)$ is expressed as follows:
\begin{eqnarray}
\mathrm{tr}(D_\sigma)=\sigma^2\sum_{i=1}^{k}\sqrt{\frac{4}{\sigma^4}\lambda_i+1}.
\end{eqnarray}
Here ${\rm det}(D_\sigma+\sigma^2I)$ can be solved as follows:
\begin{eqnarray}
{\rm det}(D_\sigma+\sigma^2I)&=&\sigma^{2l}{\rm det}\left(\frac{D_\sigma}{\sigma^2}+I\right)\\\nonumber&=&\sigma^{2l}\left(\sqrt{\frac{4}{\sigma^4}\lambda_1+1}+1\right)\left(\sqrt{\frac{4}{\sigma^4}\lambda_2+1}+1\right)\cdots\left(\sqrt{\frac{4}{\sigma^4}\lambda_k+1}+1\right).
\end{eqnarray}
Therefore, ${\rm log~det}(D_\sigma+\sigma^2I)$ is
\small
\begin{eqnarray}
{\rm log~det}(D_\sigma+\sigma^2I)=2l{\rm log}\sigma+\sum_{i=1}^k {\rm log}\left(\sqrt{\frac{4}{\sigma^4}\lambda_i+1}+1\right).
\end{eqnarray}
Taken together, 
\begin{eqnarray}
\mathcal{F}&=&{\rm tr}(D_\sigma)-l\sigma^2\left(1-{\rm log}(2\sigma^2)\right)-\sigma^2{\rm log~det}\left(D_\sigma+\sigma^2I\right)\\\nonumber
&=&\sigma^2\sum_{i=1}^{k}\sqrt{\frac{4}{\sigma^4}\lambda_i+1}-l\sigma^2\left(1-{\rm log}(2\sigma^2)\right)-\sigma^2\left(2l{\rm log}\sigma+\sum_{i=1}^k {\rm log}\left(\sqrt{\frac{4}{\sigma^4}\lambda_i+1}+1\right)\right)\\\nonumber
&=&\sigma^2\sum_{i=1}^{k}\sqrt{\frac{4}{\sigma^4}\lambda_i+1}-\sigma^2\sum_{i=1}^k {\rm log}\left(\sqrt{\frac{4}{\sigma^4}\lambda_i+1}+1\right)-l\sigma^2(1-{\rm log2})\\\nonumber
&=&\sigma^2\left(\sum_{i=1}^{k}\sqrt{\frac{4}{\sigma^4}\lambda_i+1}-\sum_{i=1}^k {\rm log}\left(\sqrt{\frac{4}{\sigma^4}\lambda_i+1}+1\right)-l{\rm log5}\right).
\end{eqnarray}
Interestingly, we found that the two solutions are the same with $2\sigma^2=\epsilon.$
\vspace{4mm}

\begin{table*}[ht]
\caption{$d(\cdot, \cdot)$ with $\gamma=1$ in a RKHS (A1) between dataset 1 and dataset 2, (A2) between dataset 1 and dataset 3, and (A3) between dataset 2  and dataset 3 using five different probability combinations \{(0.1,0.9), (0.3,0.7), (0.5,0.5), (0.7,0.3), (0.9,0.1)\} for two Gaussian components of each Gaussian mixture.  }
  \centering
\begin{tabular}{ |c|c|c|c|c|c|  }
\hline
\multicolumn{6}{|c|}{$\gamma=1$} \\[0.1em]
\hline
\multicolumn{6}{|c|}{(A1)~~~~~ Dataset 1 [row: $(p_0^1,p_0^2)$] vs Dataset 2 [column: $(p_1^1,p_1^2)$]} \\[0.1em]
\hline
$(p_0^1,p_0^2)/(p_1^1,p_1^2)$	&(0.1, 0.9)	&(0.3, 0.7)	&(0.5, 0.5)	&
(0.7, 0.3)	&(0.9, 0.1) \\
\hline
(0.1, 0.9)	&1.292	&1.249	&1.206	&1.164	&1.121\\
(0.3, 0.7)	&1.293	&1.246	&1.203	&1.160	&1.118\\
(0.5, 0.5)	&1.294	&1.247	&1.200	&1.157	&1.114\\
(0.7, 0.3)	&1.295	&1.248	&1.201	&1.154	&1.111\\
(0.9, 0.1)	&1.295	&1.248	&1.201	&1.155	&1.108\\
\hline
\multicolumn{6}{|c|}{(A2)~~~~~ Dataset 1 [row: $(p_0^1,p_0^2)$] vs Dataset 3 [column: $(p_2^1,p_2^2)$]} \\[0.1em]
\hline
$(p_0^1,p_0^2)/(p_2^1,p_2^2)$	&(0.1, 0.9)	&(0.3, 0.7)	&(0.5, 0.5)	&
(0.7, 0.3)	&(0.9, 0.1) \\
\hline
(0.1, 0.9)	&1.185	&1.123	&1.060	&0.998	&0.935\\
(0.3, 0.7)	&1.289	&1.227	&1.164	&1.102	&1.091\\
(0.5, 0.5)	&1.394	&1.331	&1.268	&1.258	&1.247\\
(0.7, 0.3)	&1.498	&1.435	&1.425	&1.414	&1.404\\
(0.9, 0.1)	&1.602	&1.591	&1.581	&1.570	&1.560\\
\hline
\multicolumn{6}{|c|}{(A3)~~~~~ Dataset 2 [row: $(p_1^1,p_1^2)$] vs Dataset 3 [column: $(p_2^1,p_2^2)$]} \\[0.1em]
\hline
$(p_1^1,p_1^2)/(p_2^1,p_2^2)$	&(0.1, 0.9)	&(0.3, 0.7)	&(0.5, 0.5)	&
(0.7, 0.3)	&(0.9, 0.1) \\
\hline
(0.1, 0.9)	&0.890	&0.844	&0.798	&0.753	&0.707\\
(0.3, 0.7)	&0.905	&0.796	&0.751	&0.705	&0.659\\
(0.5, 0.5)	&0.921	&0.812	&0.703	&0.657	&0.612\\
(0.7, 0.3)	&0.936	&0.827	&0.718	&0.609	&0.564\\
(0.9, 0.1)	&0.952	&0.843	&0.734	&0.625	&0.516\\
\hline
\end{tabular}
\end{table*}

\begin{table*}[ht]
\caption{$d(\cdot, \cdot)$ with $\gamma=10$ in a RKHS (B1) between dataset 1 and dataset 2, (B2) between dataset 1 and dataset 3, and (B3) between dataset 2  and dataset 3 using five different probability combinations \{(0.1,0.9), (0.3,0.7), (0.5,0.5), (0.7,0.3), (0.9,0.1)\} for two Gaussian components of each Gaussian mixture.  }
  \centering
\begin{tabular}{ |c|c|c|c|c|c|  }
\hline
\multicolumn{6}{|c|}{$\gamma=10$} \\[0.1em]
\hline
\multicolumn{6}{|c|}{(B1)~~~~~ Dataset 1 [row: $(p_0^1,p_0^2)$] vs Dataset 2 [column: $(p_1^1,p_1^2)$]} \\[0.1em]
\hline
$(p_0^1,p_0^2)/(p_1^1,p_1^2)$	&(0.1, 0.9)	&(0.3, 0.7)	&(0.5, 0.5)	&
(0.7, 0.3)	&(0.9, 0.1) \\
\hline
(0.1, 0.9)	&1.661	&1.618	&1.575	&1.532	&1.489\\
(0.3, 0.7)	&1.666	&1.607	&1.564	&1.521	&1.478\\
(0.5, 0.5)	&1.670	&1.611	&1.552	&1.509	&1.466\\
(0.7, 0.3)	&1.675	&1.616	&1.557	&1.498	&1.455\\
(0.9, 0.1)	&1.680	&1.621	&1.562	&1.503	&1.443\\
\hline
\multicolumn{6}{|c|}{(B2)~~~~~ Dataset 1 [row: $(p_0^1,p_0^2)$] vs Dataset 3 [column: $(p_2^1,p_2^2)$]} \\[0.1em]
\hline
$(p_0^1,p_0^2)/(p_2^1,p_2^2)$	&(0.1, 0.9)	&(0.3, 0.7)	&(0.5, 0.5)	&
(0.7, 0.3)	&(0.9, 0.1)\\
\hline
(0.1, 0.9)	&1.692	&1.610	&1.528	&1.446	&1.364\\
(0.3, 0.7)	&1.742	&1.660	&1.578	&1.496	&1.480\\
(0.5, 0.5)	&1.792	&1.710	&1.627	&1.612	&1.596\\
(0.7, 0.3)	&1.841	&1.759	&1.743	&1.728	&1.712\\
(0.9, 0.1)	&1.891	&1.875	&1.859	&1.843	&1.828\\
\hline
\multicolumn{6}{|c|}{(B3)~~~~~ Dataset 2 [row: $(p_1^1,p_1^2)$] vs Dataset 3 [column: $(p_2^1,p_2^2)$]} \\[0.1em]
\hline
$(p_1^1,p_1^2)/(p_2^1,p_2^2)$	&(0.1, 0.9)	&(0.3, 0.7)	&(0.5, 0.5)	&
(0.7, 0.3)	&(0.9, 0.1)\\
\hline
(0.1, 0.9)	&1.282	&1.371	&1.461	&1.550	&1.639\\
(0.3, 0.7)	&1.355	&1.124	&1.214	&1.303	&1.392\\
(0.5, 0.5)	&1.427	&1.197	&0.967	&1.056	&1.145\\
(0.7, 0.3)	&1.500	&1.270	&1.039	&0.809	&0.898\\
(0.9, 0.1)	&1.573	&1.342	&1.112	&0.882	&0.651\\
\hline
\end{tabular}
\end{table*}

\end{document}